\documentclass[colorlinks=true, linkcolor=black, urlcolor=blue, citecolor=blue, letterpaper, 10 pt, conference]{ieeeconf}  

\IEEEoverridecommandlockouts                              

\usepackage{color}
\usepackage{array}

\usepackage{graphicx} 

\usepackage{mathptmx} 
\usepackage{times} 
\usepackage{amsmath} 
\usepackage{amssymb}  
\usepackage{color}
\usepackage{bm}
\usepackage{rotating}
\usepackage{subfigure}
\usepackage{tabularx}
\usepackage{colortbl}
\usepackage{hhline}
\usepackage{multirow}
\usepackage{verbatim}
\usepackage{cite}
\usepackage{siunitx}
\usepackage{graphicx}
\usepackage{duckuments}
\usepackage{soul,xcolor}
\usepackage{duckuments}
\usepackage{siunitx}
\setstcolor{red}
\usepackage[ruled, lined, linesnumbered, commentsnumbered, longend]{algorithm2e}
\usepackage{algpseudocode}
\usepackage[capitalise,noabbrev]{cleveref}
\usepackage{amsfonts}
\usepackage{siunitx}
\usepackage{booktabs}

\usepackage{multirow}
\usepackage[flushleft]{threeparttable}
\usepackage[acronym]{glossaries}
\usepackage{float}
\glsdisablehyper

\newacronym{dof}{DoF}{Degree of Freedoms}
\newacronym{tof}{ToF}{Time-of-Flight}
\newacronym{hri}{HRI}{Human Robot Interaction}
\newacronym{fov}{FoV}{Field of View}
\newacronym{pc}{PC}{Point Cloud}

\newcommand{\TODO}[1]{\textcolor{red}{TODO: #1}}
\newcommand{\ale}[1]{\textcolor{black}{#1}}
\newcommand{\gi}[1]{\textcolor{black}{#1}}

\newcommand{\jack}[1]{\textcolor{green}{Jack: #1}}

\begin{document}

\title{\LARGE \bf
Estimating Scene Flow in Robot Surroundings with Distributed Miniaturised Time-of-Flight Sensors 
}

\author{Jack Sander$^{\dag}$, Giammarco Caroleo$^{\dag *}$, Alessandro Albini and Perla Maiolino
	\thanks{Authors are with the Oxford Robotics Institute (ORI), University of Oxford, UK.}
	\thanks{$^{*}$ Corresponding author.}
    \thanks{$^{\dag}$ Authors equally contributed to the paper.}
	\thanks{This work was supported by the SESTOSENSO project (HORIZON EUROPE Research and Innovation Actions under GA number 101070310).}
}

\maketitle


\newpage

\begin{abstract}

\ale{Tracking} the motion of humans or objects in a robot's surroundings is essential to improve safe robot motions and reactions. 
In this work, we present an approach for scene flow estimation from 
low-density and noisy point clouds acquired from miniaturised \gls{tof} sensors distributed across the robot's body. 
The proposed method clusters points from consecutive frames and applies the Iterative Closest Point (ICP) algorithm to estimate a dense motion flow, with additional steps introduced to mitigate the impact of sensor noise and low-density data points. Specifically, we employ a fitness-based classification to distinguish between stationary and moving points and an inlier removal strategy to refine geometric correspondences. 

The proposed approach is validated in an experimental setup \ale{where 24 \gls{tof} are used to estimate the velocity of an object moving at different controlled speeds}.
Experimental results show that the method \gi{consistently approximates the direction of the motion and its magnitude with an error which is in line with sensor noise.} 


\end{abstract}

\glsresetall
\section{Introduction}\label{sec:intro}

Robots operating in cluttered or shared environments must be aware of their surroundings to plan safe motions effectively. Tracking the motion of nearby humans and obstacles is crucial for detecting and reacting to potential collisions, as well as improving human-robot collaboration~\cite{lasota2015analyzing, smith2023socially, bellotto_2009, liu2017human}.

\ale{Motion tracking is commonly addressed in the literature through} \textit{scene flow} estimation, which represents the three-dimensional motion of visible points in a scene across consecutive time frames \cite{vedula_2005,menze2018object,guizilini_2022}.
This problem has been extensively studied, especially in autonomous driving \cite{guizilini_2022,vogel2013piecewise,jiang2019sense,liu2019flownet3d,menze2015object} and object manipulation \cite{shao_2018,duisterhof2023deformgs,xu2020learning}. 
%
Past literature on motion tracking is based on developing both model-based and data-driven algorithms \cite{zhai2021optical}, relying on high-density environmental representations in the form of point clouds acquired via high-resolution depth cameras or LiDAR systems~\cite{li2023deep,muthu2023survey}. 
%
%
Despite the reliability of these sensors, when considering robotic manipulators, their integration poses several limitations. Cameras are typically mounted on the robot's end-effector, which restricts the \gls{fov} and reduces spatial awareness. Furthermore, integrating multiple depth cameras across different robot links is technically challenging due to their size. Cameras placed in the environment can mitigate this issue; however, bandwidth requirements must be considered \cite{ding2019stitching}, and they remain prone to occlusions, especially in cluttered settings, while also tethering the robot to external sensing units. Similarly, LiDAR systems face scalability issues, as they are bulky, expensive, and impractical to distribute across the robot body.

Recent advances in miniaturised \gls{tof} sensors have made them an attractive alternative. These sensors are lightweight, cost-effective, and energy-efficient. Unlike LiDAR or depth camera systems, they can be distributed across the robot body, enabling more flexible, untethered perception while reducing occlusions \cite{ding_2020, tsuji2019proximity}.
\ale{To date, their usage in robotics has shown promise} in obstacle and human avoidance \cite{kumar2019speed,ding_2019, tsuji2019proximity,ding_2020,proxy-tactile}, gesture recognition for human-robot interaction \cite{anil_2020}, manipulation \cite{yang_2017,sasaki_2018}, as well as localisation and pose estimation \cite{ruget2022pixels2pose, caroleo2025tiny, caroleo2025soft}.
\ale{However, despite their compact size and low weight}, the main drawbacks of miniaturised \gls{tof} sensors are their low spatial resolution\footnote{Most recent models provide an $8\times8$ depth measurements matrix that can be converted into a point cloud.} and relatively high noise compared to traditional cameras and LiDAR, making their measurements less reliable. 
%


\begin{figure}[t]
	\centering
	\includegraphics[width=0.9\linewidth]{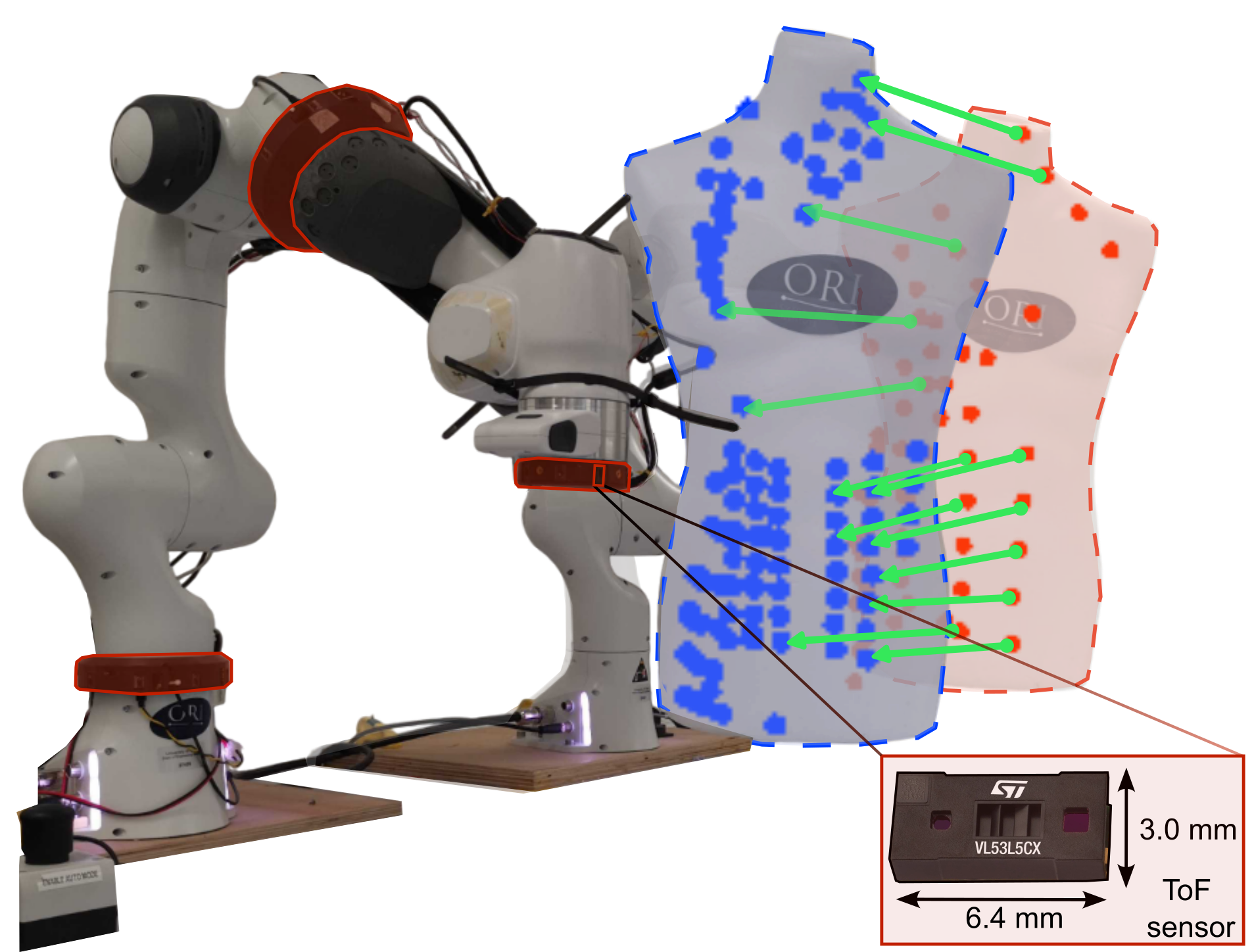}
	\caption{The flow estimation of a mannequin moving from right to left towards a robot is shown. The mannequin, mounted on a robot arm (right), is captured by distributed ToF sensors on the other robot (left). Point clouds from two time instants are coloured red and blue, with green arrows indicating predicted flow. Red shaded areas highlight the proximity sensors on the robot, and a zoom-in (bottom right) shows the sensor image and dimensions. }
	\label{fig:intro}
\end{figure}

\ale{This paper wants to investigate the use of distributed miniaturised \gls{tof} for dense scene flow estimation, which is made challenging by the sparsity, low density, and noise of the acquired data.} 
\ale{An illustration of the addressed problem is shown in
\cref{fig:intro}, where} a robot equipped with distributed \gls{tof} on three links (base, elbow, and end-effector), is used to estimate the dense scene flow (represented by green arrows) of an approaching mannequin \ale{operated with controlled motion speed.}

The proposed approach \ale{is based on the family of algorithms that leverage} clustering and Iterative Closest Point (ICP)~\cite{besl1992method} to establish geometric correspondences between point clouds acquired at different time frames, as in~\cite{ICP-Flow,li2022rigidflow}.
\ale{This approach was selected over data-driven alternatives because it does not require large labelled datasets, which would be costly and time-consuming to collect.}
\ale{Although fine-tuning a pretrained model on a smaller dataset could be a viable option, most existing models are trained on data from LiDAR or high-resolution depth cameras. Consequently, they are unlikely to generalise well to our setting due to significant differences in data density, resolution, and noise.} 
%

%
\ale{To address the aforementioned challenges, we introduce two additional steps to the pipeline based on ICP geometrical matching:} (i) the fitness score is leveraged to distinguish between actual motion and noise-induced displacements; (ii)~an inlier removal step is applied to discard points with low fitness after ICP computation.
%
%
This last step is generally unnecessary in state-of-the-art methods~\cite{ICP-Flow,li2022rigidflow}, as they rely on high-density, low-noise data. 
However, in our case, it is crucial as the application of ICP may not always converge to find a good geometric matching.
\ale{The proposed approach is validated using the experimental setup shown in \cref{fig:intro}, where it is used to estimate the mannequin's speed across various experimental configurations. In addition, we compare our method to a baseline based solely on ICP to demonstrate how the introduced modifications help address the challenges associated with \gls{tof} data, as previously discussed.}

The paper is organized as follows: \cref{sec:methodology} introduces an overview of the sensing technology and explains the proposed method. Validation experiments and setup are described in \cref{sec:validation}. \cref{sec:results} presents and discusses the results. Conclusions follow.

\section{Methodology}
\label{sec:methodology}

This Section describes the proposed approach. First, it provides an overview of the miniaturised \gls{tof} technology and the type of data it produces. Then, it presents a description of the algorithm that estimates the scene flow.

\subsection{Point Cloud Representation from Distributed ToF Sensors}

\begin{figure}[t]
	\centering
    \includegraphics[width=0.9\linewidth]{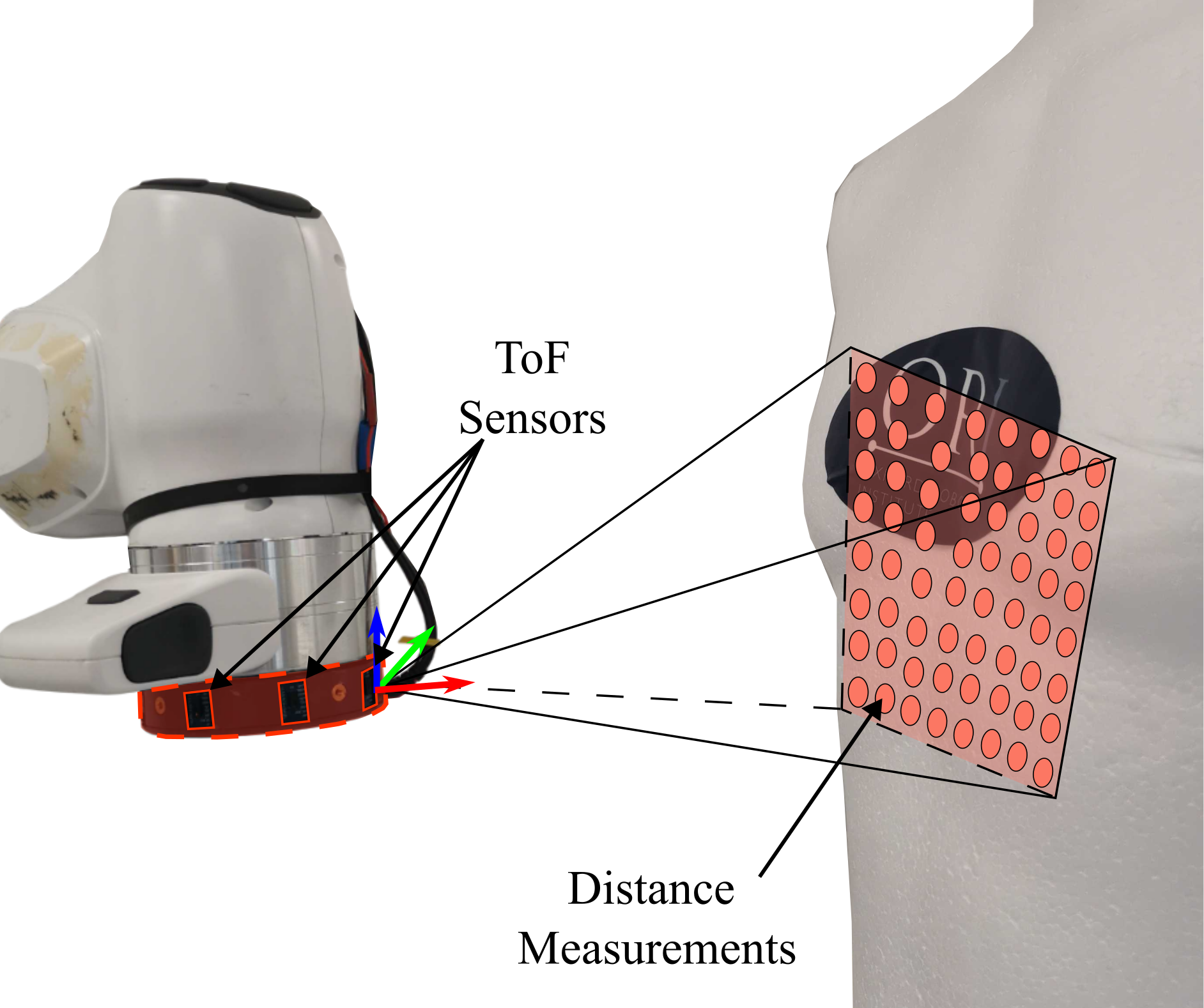}
	\caption{The robot's end-effector equipped with distributed \gls{tof} sensors is shown. A visualization of the squared \gls{fov}, the measured point cloud and the origin with the reference frame are also illustrated for one of the sensors. }
	\label{fig:tof_out}
\end{figure}

The robot is assumed to be equipped with a number of independent \gls{tof} sensors.
At each discrete time instant $k$, each \gls{tof} sensor provides a set of measurements representing the absolute distance from its origin to objects within its \gls{fov} (see \cref{fig:tof_out}).

Distance measurements collected at the $k$-th time step can be converted to a low-resolution point cloud using the sensor’s intrinsic parameters, typically available in its datasheet \cite{ToFdatasheet}. 
Assuming the position of each \gls{tof} is known with respect to the robot base\footnote{This requires the single \gls{tof} to be spatially calibrated. As sensors are integrated into custom plastic mounts, their positions and orientation can be retrieved from the mounts' CAD models.},  the individual point clouds can be transformed and merged into a common reference frame. 
The resulting point cloud $P_k$ consists of a sparse set of points approximating the robot's surroundings. 
Notably, since the \gls{tof} sensors are distributed across the robot body, their measurements may also include points belonging to the robot itself. To isolate only the environment, the robot’s shape is filtered out using a ray-casting technique similar to that in~\cite{kumar2019speed}.

\ale{
As discussed in the Introduction, the point cloud generated from \gls{tof} distance measurements has significantly lower resolution and reliability compared to traditional sensors used for scene flow estimation. 
For instance, the individual \gls{tof} sensor used in this work (details in \cref{sec:validation}) provides 64 distance measurements, with an error ranging from 8\% to 11\% over a distance range of \SI{0.2}{\meter} to \SI{4}{\meter}.
This poses a considerable challenge for ICP-based approaches, as geometric matching between successive point clouds becomes difficult due to the low spatial resolution and high uncertainty of the measurements. The following section outlines the approach we propose to address these challenges.
}

\subsection{Scene Flow Estimation}

The algorithm estimates the scene flow from two point clouds $P_k$ and $P_{k-n}$ collected at two different time instants, \ale{ where $k \in \mathbb{Z}$, $n \in \mathbb{Z}^+$, with $k$ representing the current discrete time step. }
Typically, scene flow is estimated between two consecutive time frames, i.e., $n=1$. 
However, in our settings, especially when considering low speeds 
- for a given sensor's sampling time $\Delta T$ -
measurement errors cause displacements larger than those induced by the actual motion. The impact of $n$ on the estimation is analysed in \cref{sec:results}.

As previously discussed, the proposed approach, detailed in \cref{alg:flow_est}, leverages ICP to establish geometric correspondences among clustered points, following a methodology similar to \cite{ICP-Flow,li2022rigidflow}. Additional steps have been introduced to address the challenges posed by \gls{tof} data. With respect to \cref{alg:flow_est}, these can be found at:
\begin{itemize}
	\item Lines \ref{alg:line:if_fit}-\ref{alg:line:endif_fit}: The fitness score is used to distinguish between stationary and non-stationary points. Additionally, displacement is compared with a noise threshold to determine if it results from actual motion or sensor noise.
	\item Lines \ref{alg:line:if_inliers}-\ref{alg:line:endif_inliers}: If the fitness score is too low, the corresponding inliers are removed from the cluster.
\end{itemize}	
 
\begin{figure}[h]
    \centering
    \includegraphics[width=0.95\linewidth]{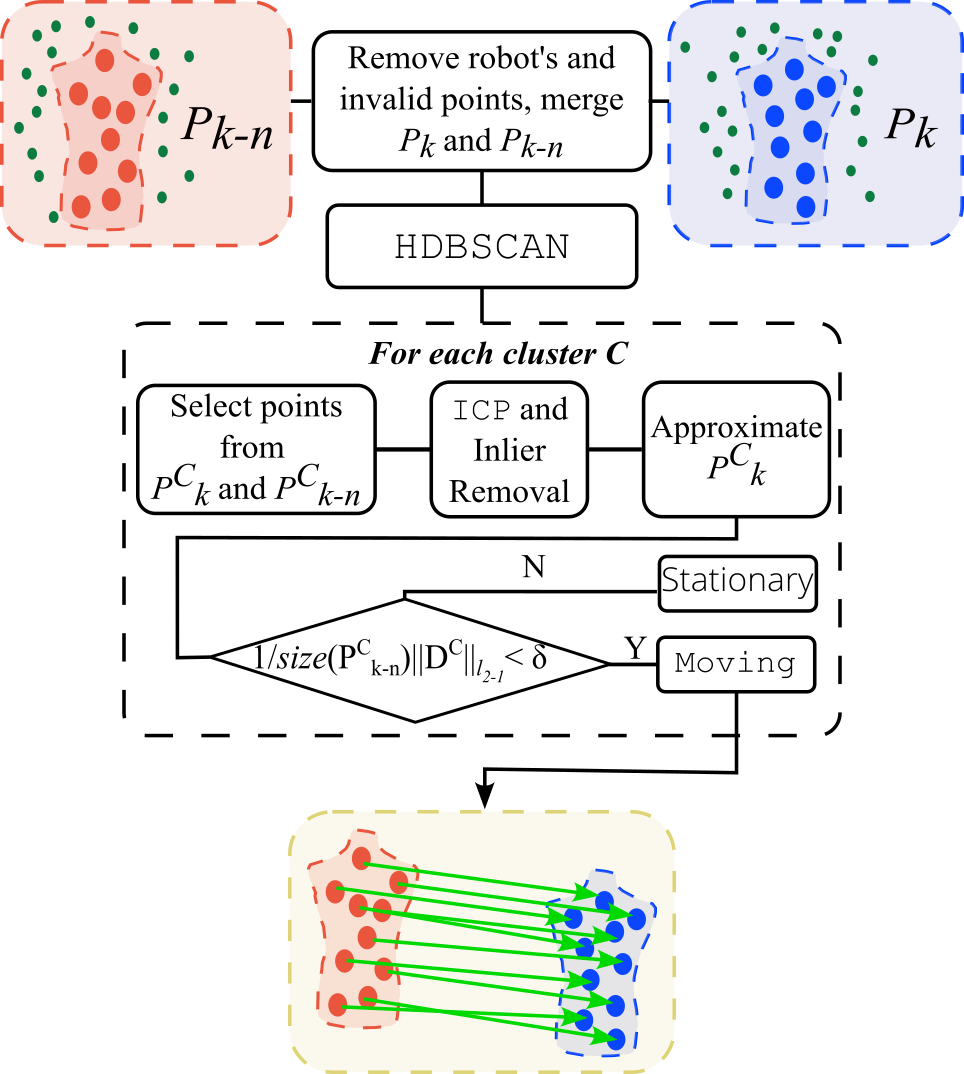}
    \caption{Flowchart representing the steps of \cref{alg:flow_est}. The two input point clouds are merged and clustered with \texttt{HDBSCAN}. For each cluster, points are processed with \texttt{ICP} and the enhancement steps devised to approximate the transformation between the two input point clouds while considering sensor measurement noise. The cluster is then flagged as \texttt{Stationary} or \texttt{Moving}. In the latter case, scene flow is estimated.}
    \label{fig:flowchart}
\end{figure}

More specifically, \cref{alg:flow_est} works as follows.
We begin by merging the two point clouds collected at time instants $k$ and $k-n$ into a single point cloud $P$. Next, clustering is performed on $P$ using \texttt{HDBSCAN} \cite{HDBSCAN}.
For each cluster $C$ \ale{(computed at \cref{alg:line:clustering})}, we extract the subset of points belonging to $P_k$ and $P_{k-n}$, denoted as $P_k^C$ and $P_{k-n}^C$ in \cref{alg:flow_est} (see \cref{alg:line:extraction}). 
If the cluster contains fewer points than an \ale{experimentally tuned} threshold $N_{min}$,  its points are considered noise, and the loop proceeds to the next cluster (see \cref{alg:line:if_npoints}). 
At Lines \ref{alg:line:goto} and \ref{alg:line:delta}, ICP is applied to compute the transformation matrix $T_C \in \Re^{4\times 4}$ between \ale{$P_{k-n}$ and $P_k$, along with the corresponding fitness score. }
\ale{The transformation matrix is applied to $P_{k-n}^C$, to compute $\hat{P}_k^C$, i.e. the estimated position of the points at the $k$-th time step}.
%
At 
\cref{alg:line:dist_matrix} it is computed $\mathbf{D}^C \in \Re^{3\times s}$, with $s=size(P_{k-n}^C)$, that is the distance between the transformed points and the initial ones. 
%
Then, the fitness score is compared against a threshold $\gamma$. If it exceeds $\gamma$, the $\ell_{2,1}$-norm of $\mathbf{D}^C$ is divided by the number of points in $P_{k-n}^C$ to get the average norm of the distance between $\hat{P}_k^C$ and $P_{k-n}^C$ points. 
\ale{The result is further compared to another threshold $\delta$ (see \cref{alg:line:dist}).} If the average displacement is small, the cluster $C$ is classified as \texttt{Stationary}. 
%
Otherwise, it is labeled as \texttt{Moving} and the velocity $\mathbf{v_i}$ of each point in $P_{k-n}^C$ is estimated (see \cref{alg:line:vp_est}). 
\\
Conversely, if the fitness score is below $\gamma$, inliers are removed from $C$, and the algorithm proceeds to \cref{alg:line:goto} to recompute ICP on the new cluster. This allows us to remove from the cluster pairs of points that are close and discriminate if the remaining ones are far from each other because of sensor noise or displacement. 
\cref{alg:flow_est} ultimately assigns a label to each cluster identified by \texttt{HDBSCAN}, categorizing its points as either \texttt{Stationary} or \texttt{Moving}. 
\Cref{fig:flowchart} details the proposed algorithm in a flowchart. 
\vspace{1mm}  
\begin{algorithm}
	\caption{Scene flow estimation}
	\label{alg:flow_est}
	\KwIn{Point clouds: $P_k$ and $P_{k-n}$. Thresholds: $\gamma$ and $\delta$. Sampling time: $\Delta T$ }
	\KwOut{Clusters classification (\texttt{Stationary}, \texttt{Moving}) and corresponding scene-flow}
	
	Merge point clouds: $P \gets P_k \cup P_{k-n}$\;
	Perform clustering: $\mathcal{C} \gets \texttt{HDBSCAN}(P)$\;  \label{alg:line:clustering}
	
	\ForEach{cluster $C \in \mathcal{C}$}{
		Extract $P_k^C$ and $P_{k-n}^C$\; \label{alg:line:extraction}
		\If{ \label{alg:line:if_npoints}
			\textit{size}$(P_k^C) < N_{min}$ \textit{AND} \textit{size}$(P_{k-n}^C) < N_{min}$ 
			} { \label{alg:line:if_size}
			\textbf{continue} \Comment{Skip to the next cluster} 
		}  \label{alg:line:endif_size}
		
		Apply ICP: $\{T_C, \text{fitness}\} \gets \text{ICP}(P_k^C, P_{k-n}^C)$\; \label{alg:line:goto}
		
		
            $\mathbf{t}^C = \texttt{GetTranslation}(T)$ \label{alg:line:delta}\;
		
		$\mathbf{R}^C = \texttt{GetRotationMatrix}(T)$ \label{alg:line:ang}\; 
            $\hat{P}_k^C = \mathbf{R}^C P_{k-n}^C + \mathbf{t}^C$ \Comment{Approximate $P_k^C$}\
            
            $\mathbf{D}^C = \hat{P}_k^C -P_{k-n}^C$\;  \label{alg:line:dist_matrix}
		\If{$\text{fitness} > \gamma$} { \label{alg:line:if_fit} 
			\If{$1/size(P_{k-n}^C)\| \mathbf{D}^C\|_{\ell_{2,1}} < \delta$} {\label{alg:line:dist}
                    Flag $C$ as \texttt{Stationary}\;
			}
				 
				 
        \Else{
				Flag $C$ as \texttt{Moving}\;
                    $\mathbf{V} = \mathbf{D}^C/(n\Delta T)$ \Comment{$\mathbf{V} = \{ \mathbf{v}_i~|~\mathbf{v}_i \in \Re^{3} \} $}
                     \label{alg:line:vp_est}
			}
		}  \label{alg:line:endif_fit}
		\Else{ \label{alg:line:if_inliers}
			Remove inliers from $C$ and go to \textbf{\cref{alg:line:goto}}\;
		} \label{alg:line:endif_inliers}
	}
    \end{algorithm}
\section{Validation}
\label{sec:validation}

\begin{figure}[b!]
	\centering
	\includegraphics[width=\linewidth]{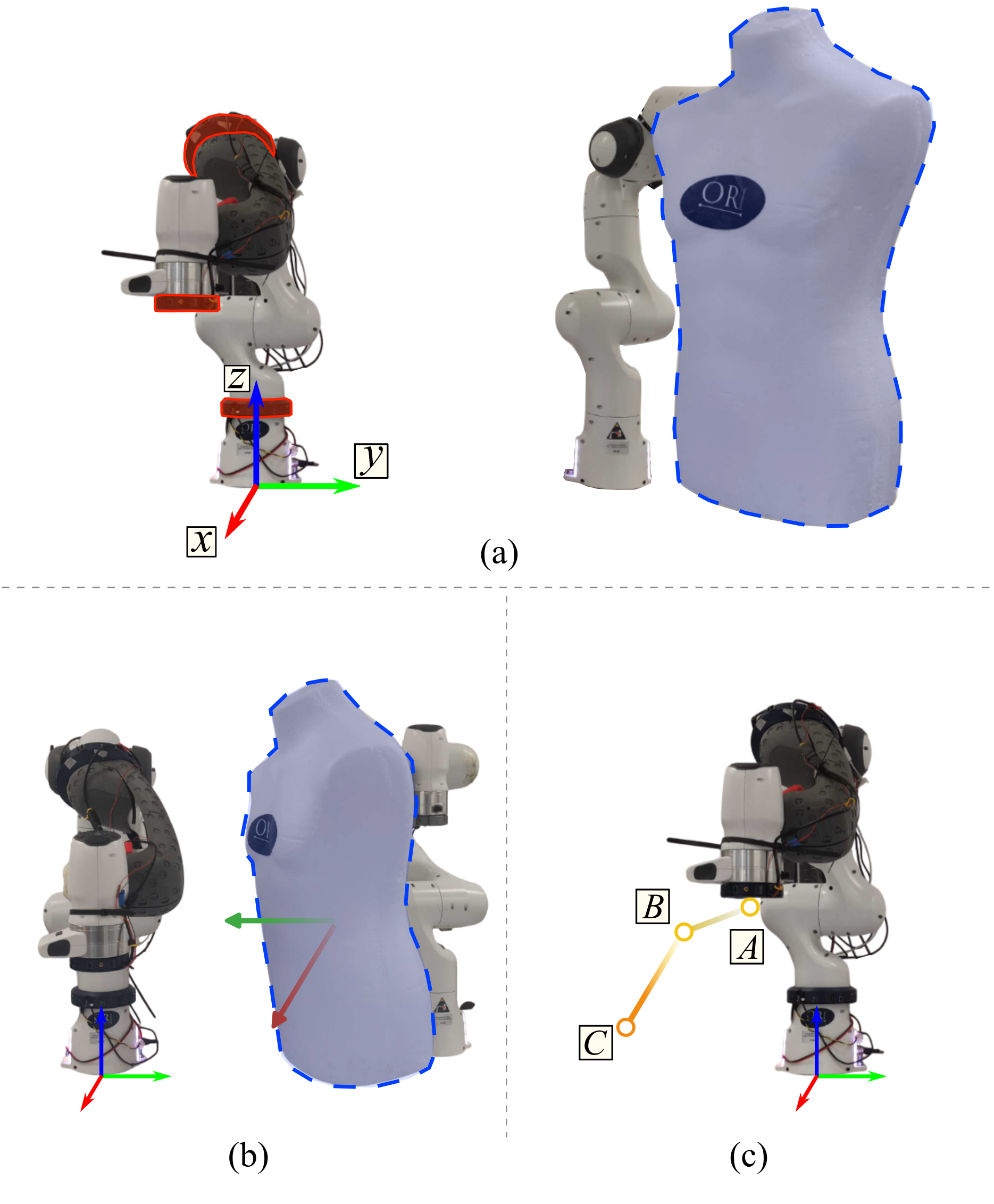}
	\caption{The experimental setup and validation scenarios are shown.
(a) Two robotic manipulators: the left one equipped with 3 rings of ToF sensors (red shaded areas), and the right one with a mannequin rigidly attached (in blue). The world frame, aligned with the ToF-equipped robot’s base, is displayed. (b)  Mannequin motions: the green arrow indicates the mannequin moving towards the sensorised robot; the red arrow shows sideways motion. (c) The commanded path of the ToF-equipped robot in one validation scenario, with labeled waypoints.}
	\label{fig:setup}
\end{figure}
\subsection{Experimental Setup}
The approach described in \cref{sec:methodology} has been validated using the experimental setup shown in \cref{fig:setup}, which consists of two Franka Emika Panda robots.
The left robot is equipped with distributed \gls{tof} sensors. Specifically, these sensors are mounted in custom 3D-printed plastic mounts, highlighted in red in \cref{fig:setup}(a). Each mount houses eight miniaturized \gls{tof} sensors, arranged in a circular configuration with equal spacing.
Each individual \gls{tof} sensor is a VL53L5CX, an off-the-shelf component manufactured by STMicroelectronics featuring a squared \gls{fov} with a diagonal angle of $65^\circ$ and a detection range between \SI{0.02}{\meter} and \SI{4}{\meter}. \ale{Distance measurements are provided in the form of an $8\times 8$ depth map.} Sensors are interconnected in a network via custom hardware and firmware, ensuring synchronized measurements at \SI{15}{\hertz}, corresponding to a sampling interval of $\Delta T \approx $~\SI{0.067}{\second}.

The second robot is \SI{0.75}{\meter} far from the first and equipped with a mannequin rigidly attached to its end-effector. This setup allows controlled movement of the mannequin at different, measurable speeds we use as ground truth.

\subsection{Experiments Description}

The proposed method is validated in three different scenarios.
\ale{Given the wide variety of possible motion trajectories for the mannequin, we focused on two cases: motion along the $y$-axis (towards the robot) and motion along the $x$-axis (tangential to the robot). These scenarios are particularly relevant for evaluating the method’s effectiveness in detecting approaching objects -- crucial for obstacle avoidance -- and in handling lateral motion, where tracking consistency becomes more challenging due to the sparsity in the measurements and their low resolution.}
In the first set of experiments, the mannequin is commanded to move towards the sensorised robot along a straight line, i.e., along the negative $y$-axis for \SI{0.3}{m}. The motion direction is indicated by the green arrow in \cref{fig:setup}(b). 
\\
In the second set, the mannequin moves sideways relative to the robot, i.e., along its $x$-axis, for \SI{0.25}{\meter}.
The third scenario analyses the case where both the robot and the mannequin move relative to each other. In this respect, the mannequin follows the same trajectory as in the first set of experiments, while the sensorised robot is commanded to follow a predefined motion: the end-effector first moves diagonally, for \SI{0.15}{\meter} along the negative $y$-axis and for \SI{0.05}{\meter} along the positive $x$-axis and then continues for other \SI{0.10}{\meter} in the same direction, as shown in \cref{fig:setup}(c), at an average speed of \SI{0.25}{\meter\per\second}.

Each experiment was repeated five times at four different target speeds: $\lbrace 0.15, 0.20, 0.25, 0.30\rbrace ~\si{\meter\per\second}$.
%
These velocities correspond to the commanded motion of the robot's flange to which the mannequin is attached. Since the motion is purely translational, every point on the mannequin moves at the same velocity as the flange. The robot’s joint encoders measure this velocity, which serves as the ground truth for validating the velocities estimated with \cref{alg:flow_est}.
Throughout the mannequin’s motion, \gls{tof} data were recorded, and \cref{alg:flow_est} was evaluated offline using the collected dataset.
\ale{It is worth noting that, despite the analyses being conducted offline, the algorithm takes approximately from 0.05 to 0.06~\si{\second} on a laptop equipped with a 1.4 GHz Quad-Core Intel Core i5 CPU. The execution time is influenced by point clouds and cluster size, and ICP convergence.}

According to the sensor datasheet \cite{ToFdatasheet}, the uncertainty in the measured distances due to noise is between 8\% and 11\% of the actual distance  (within the 0.2--4 \SI{}{m} range).
%
As explained in \cref{sec:methodology}, if the mannequin moves too slowly, it becomes impossible to distinguish actual motion from noise-induced displacements when using only two consecutive time instants.
To address this issue, we compare point clouds collected at time step $k-n$. However, a downside of this approach is that it introduces a delay in the estimation, proportional to $n$. 
Given the relatively slow sampling rate of the \gls{tof} sensors (\SI{15}{\hertz}) and the mannequin’s travel distance when approaching the robot ({\SI{0.30}{\meter}), we limited the maximum target speed to \SI{0.30}{\meter\per\second} to ensure a sufficient number of samples ($n$) could be collected before the mannequin's stopped. 
\ale{The value of $n$ also affects the selection of the minimum target speed \SI{0.15}{\meter\per\second}, as going slower would require a large $n$ thus introducing a significant delay.}
The effect of $n$ on the results is analysed in \cref{sec:results}, using data from the first experimental setting -- where the robot remains stationary and the mannequin moves along the negative $y$-axis. This scenario is chosen because it represents the most critical case in real-world applications from a safety perspective, given that the object is approaching the robot.

After selecting $n$, \cref{alg:flow_est} was evaluated across the three experimental settings using two criteria.
The first criterion assesses the \textit{direction} of the scene flow by measuring how closely the computed linear velocity vectors align with the ground truth. This is particularly relevant for tasks like segmentation, where motion direction is the primary concern and velocity magnitude is less critical.
The second criterion evaluates both \textit{direction} and \textit{magnitude}, which is essential for control tasks that require accurate speed estimation to ensure safe robot motion. 
\ale{Additionally, \cref{alg:flow_est} is compared with a baseline that excludes the two steps introduced to handle the complexities in the \gls{tof} data. This comparison aims to quantify the impact of the proposed modifications on estimation accuracy.}

%

The value for the \ale{ thresholds $N_{min}$ and $\gamma$ used to compare the number of points and fitness score in \cref{alg:flow_est} were experimentally tuned to 60 and 0.7, respectively.} The threshold $\delta$ was set to \SI{0.04}{\meter}, which corresponds to the expected uncertainty in distance measurements (8-11\%) over the mannequin's travel range.

\section{Results and Discussion}
\label{sec:results}

\begin{figure}[t]
	\centering
	\includegraphics[width=\linewidth]{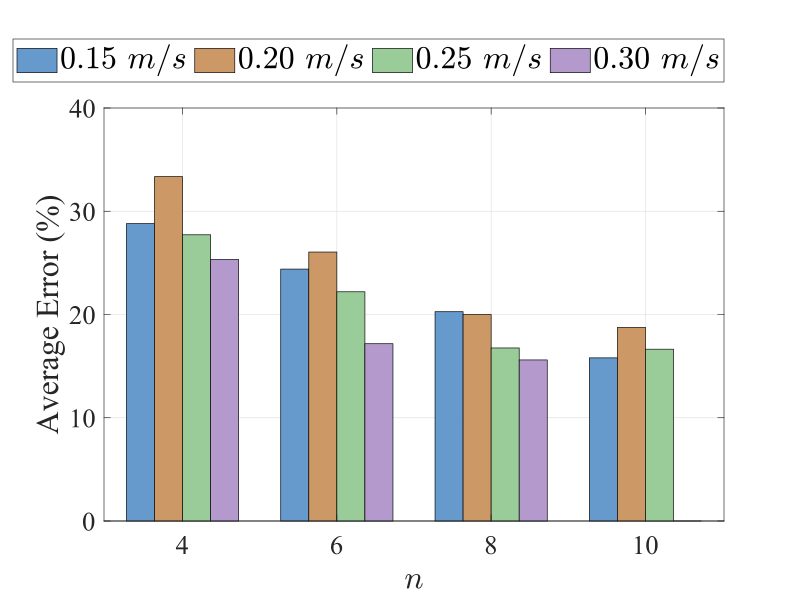}
	\caption{Analysis on the average error on the estimated speed obtained by varying $n$.}
	\label{fig:res_n}
\end{figure}

\begin{table}[]\caption{Summary of Speed, Angle Deviation, and Magnitude Statistics for the $y$-axis experiment with our method and the baseline.}
\resizebox{\columnwidth}{!}{
\begin{tabular}{|c|c|cc|cl|}
\hline
\rowcolor[HTML]{EFEFEF} 
\cellcolor[HTML]{EFEFEF} &
  \cellcolor[HTML]{EFEFEF} &
  \multicolumn{2}{c|}{\cellcolor[HTML]{EFEFEF}\textbf{\begin{tabular}[c]{@{}c@{}}Angle\\ Deviation ($^\circ$)\end{tabular}}}&
  \multicolumn{2}{c|}{\cellcolor[HTML]{EFEFEF}\textbf{\begin{tabular}[c]{@{}c@{}}Linear Velocity\\ Error (m/s)\end{tabular}}} \\ \cline{3-6} 
\rowcolor[HTML]{EFEFEF} 
\multirow{-2}{*}{\cellcolor[HTML]{EFEFEF}\textbf{Experiment}} &
  \multirow{-2}{*}{\cellcolor[HTML]{EFEFEF}\textbf{\begin{tabular}[c]{@{}c@{}}Target\\ Speed (m/s)\end{tabular}}} &
  \multicolumn{1}{c|}{\cellcolor[HTML]{EFEFEF}\textbf{Mean}} &
  \textbf{Std} &
  \multicolumn{1}{c|}{\cellcolor[HTML]{EFEFEF}\textbf{Mean}} &
  \multicolumn{1}{c|}{\cellcolor[HTML]{EFEFEF}\textbf{Std}} \\ \hline
\cellcolor[HTML]{EFEFEF} &
  0.15 &
  \multicolumn{1}{c|}{\cellcolor[HTML]{DAE8FC}9.197} &
  \cellcolor[HTML]{DAE8FC}2.119 &
  \multicolumn{1}{c|}{\cellcolor[HTML]{DAE8FC}0.030} &
  \cellcolor[HTML]{DAE8FC}0.008 \\
\cellcolor[HTML]{EFEFEF} &
  0.20 &
  \multicolumn{1}{c|}{\cellcolor[HTML]{DAE8FC}7.663} &
  \cellcolor[HTML]{DAE8FC}1.543 &
  \multicolumn{1}{c|}{\cellcolor[HTML]{DAE8FC}0.040} &
  \cellcolor[HTML]{DAE8FC}0.012 \\
\cellcolor[HTML]{EFEFEF} &
  0.25 &
  \multicolumn{1}{c|}{\cellcolor[HTML]{DAE8FC}7.842} &
  \cellcolor[HTML]{DAE8FC}1.029 &
  \multicolumn{1}{c|}{\cellcolor[HTML]{DAE8FC}0.042} &
  \cellcolor[HTML]{DAE8FC}0.009 \\
\multirow{-4}{*}{\cellcolor[HTML]{EFEFEF}\textbf{y-axis (ours)}} &
  0.30 &
  \multicolumn{1}{c|}{\cellcolor[HTML]{DAE8FC}10.313} &
  \cellcolor[HTML]{DAE8FC}2.500 &
  \multicolumn{1}{c|}{\cellcolor[HTML]{DAE8FC}0.067} &
  \cellcolor[HTML]{DAE8FC}0.013 \\ \hline
\cellcolor[HTML]{EFEFEF} &
  0.15 &
  \multicolumn{1}{l|}{77.519} &
  \multicolumn{1}{l|}{43.979} &
  \multicolumn{1}{l|}{0.125} &
  0.049 \\
\cellcolor[HTML]{EFEFEF} &
  0.20 &
  \multicolumn{1}{c|}{81.996} &
  45.324 &
  \multicolumn{1}{c|}{0.170} &
  0.064 \\
\cellcolor[HTML]{EFEFEF} &
  0.25 &
  \multicolumn{1}{c|}{84.401} &
  45.355 &
  \multicolumn{1}{c|}{0.215} &
  0.078 \\
\multirow{-4}{*}{\cellcolor[HTML]{EFEFEF}\textbf{y-axis (baseline)}} &
  0.30 &
  \multicolumn{1}{c|}{100.634} &
  30.096 &
  \multicolumn{1}{c|}{0.289} &
  0.043 \\ \hline
\end{tabular}\label{tab:speed_comparison}
}
\end{table}

\begin{table}[]\caption{Summary of Speed, Angle Deviation, and Magnitude Statistics for the $x$-axis and the $y$-axis with both robots moving experiments.}
\resizebox{\columnwidth}{!}{
\begin{tabular}{|c|c|cc|cc|}
\hline
\rowcolor[HTML]{EFEFEF} 
\cellcolor[HTML]{EFEFEF} &
  \cellcolor[HTML]{EFEFEF} &
  \multicolumn{2}{c|}{\cellcolor[HTML]{EFEFEF}\textbf{\begin{tabular}[c]{@{}c@{}}Angle\\ Deviation ($^\circ$)\end{tabular}}} &
  \multicolumn{2}{c|}{\cellcolor[HTML]{EFEFEF}\textbf{\begin{tabular}[c]{@{}c@{}}Linear Velocity\\ Error (m/s)\end{tabular}}} \\ \cline{3-6} 
\rowcolor[HTML]{EFEFEF} 
\multirow{-2}{*}{\cellcolor[HTML]{EFEFEF}\textbf{Experiment}} &
  \multirow{-2}{*}{\cellcolor[HTML]{EFEFEF}\textbf{\begin{tabular}[c]{@{}c@{}}Target\\ Speed (m/s)\end{tabular}}} &
  \multicolumn{1}{c|}{\cellcolor[HTML]{EFEFEF}\textbf{Mean}} &
  \textbf{Std} &
  \multicolumn{1}{c|}{\cellcolor[HTML]{EFEFEF}\textbf{Mean}} &
  \textbf{Std} \\ \hline
\cellcolor[HTML]{EFEFEF}                                                 & 0.15 & \multicolumn{1}{c|}{17.028} & 0.073 & \multicolumn{1}{c|}{0.066} & 0.001 \\
\cellcolor[HTML]{EFEFEF}                                                 & 0.20 & \multicolumn{1}{c|}{16.368} & 1.975 & \multicolumn{1}{c|}{0.096} & 0.007 \\
\cellcolor[HTML]{EFEFEF}                                                 & 0.25 & \multicolumn{1}{c|}{10.120} & 4.705 & \multicolumn{1}{c|}{0.120} & 0.014 \\
\multirow{-4}{*}{\cellcolor[HTML]{EFEFEF}\textbf{x-axis}}                & 0.30 & \multicolumn{1}{c|}{10.277} & 2.288 & \multicolumn{1}{c|}{0.148} & 0.017 \\ \hline
\cellcolor[HTML]{EFEFEF}                                                 & 0.15 & \multicolumn{1}{c|}{15.087} & 5.789 & \multicolumn{1}{c|}{0.048} & 0.014 \\
\cellcolor[HTML]{EFEFEF}                                                 & 0.20 & \multicolumn{1}{c|}{14.940} & 6.474 & \multicolumn{1}{c|}{0.061} & 0.02  \\
\cellcolor[HTML]{EFEFEF}                                                 & 0.25 & \multicolumn{1}{c|}{16.931} & 5.972 & \multicolumn{1}{c|}{0.085} & 0.031 \\
\multirow{-4}{*}{\cellcolor[HTML]{EFEFEF}\textbf{Robot with ToF moving}} & 0.30 & \multicolumn{1}{c|}{20.822} & 8.010 & \multicolumn{1}{c|}{0.118} & 0.034 \\ \hline
\end{tabular}}\label{tab:speed_stats_others}
\end{table}
\begin{figure}[h!]
	\centering
        \includegraphics[]{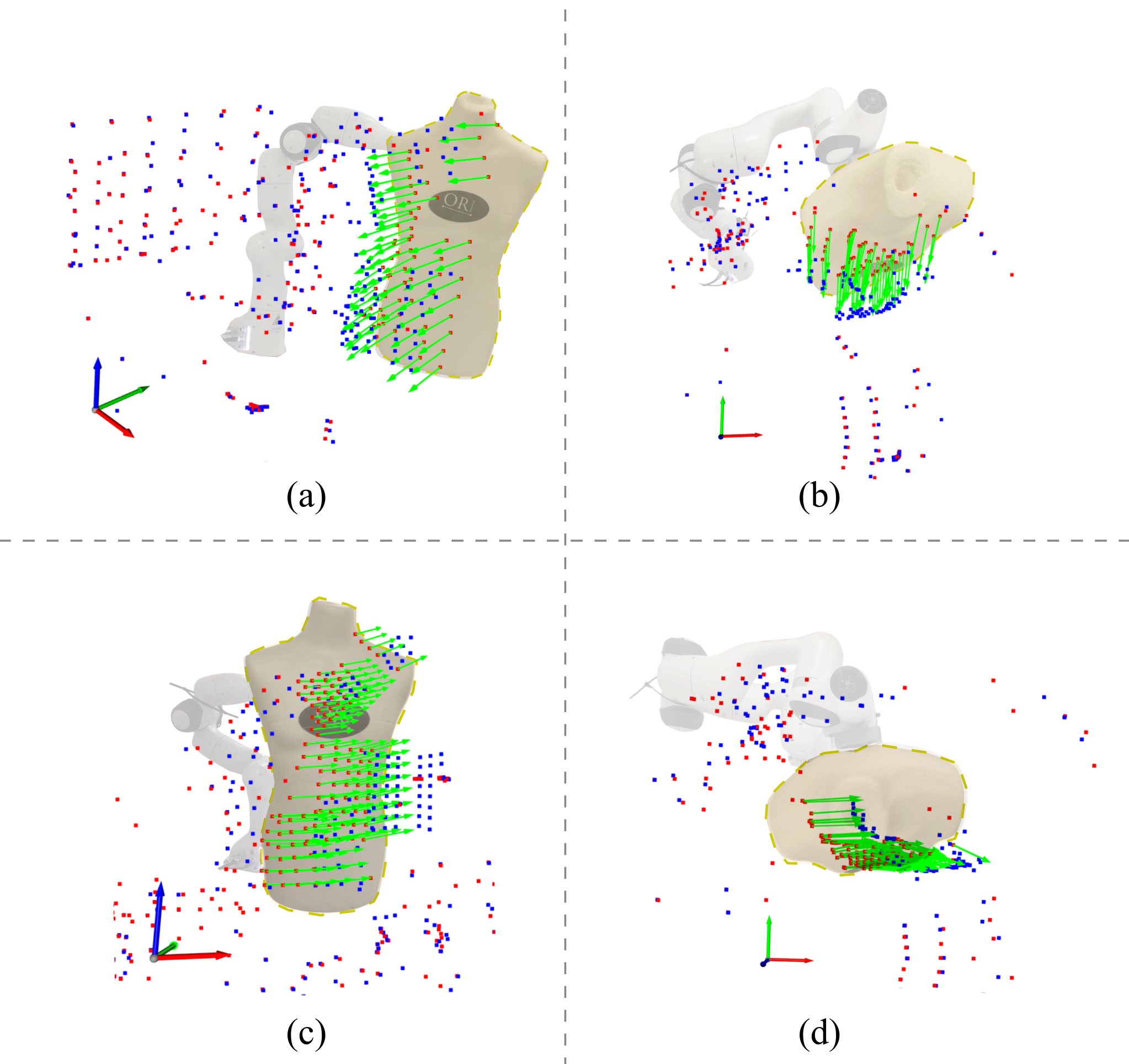}
	\caption{Examples of the predicted flow field for two validation scenarios. (a) 3D view of the flow of the mannequin approaching the robot. (b) Top view of the flow of the mannequin approaching the robot. (c) 3D view of the flow of the mannequin moving sideways. (d) Top view of the flow of the mannequin moving sideways. }
	\label{fig:res_flow}
\end{figure}

The results of the analysis on the value of $n$ are presented in the bar plot shown in \cref{fig:res_n}. The horizontal axis reports increasing values of $n$, while the vertical axis shows the average estimation error expressed as a percentage, with different colors indicating the four target speeds.
The error on the vertical axis is computed as follows: first, the linear velocity error, $\Vert \bar{\mathbf{v}}_k - \mathbf{v}_{i,k} \Vert$, is calculated at each time instant $k$ for every $i$-th point in the point cloud, where $\bar{\mathbf{v}}_{k}$ is the measured velocity applied to the mannequin\footnote{Note: Since the mannequin undergoes pure translational motion, the velocity is the same for all points on its body.}. 
This error is then averaged over time and across all points for a single trial of the experiment. Finally, the result is further averaged over the five trials and normalised by the target speed to express it as a percentage.

As shown in \cref{fig:res_n}, the average error decreases as $n$ increases. However, for $n=10$, the error for the target speed of \SI{0.3}{\meter\per\second} is not reported, as it could not be computed. This is because, at large values of $n$, the two point clouds $P_k$ and $P_{k-n}$ become too distant from each other, making it impossible for \texttt{HDBSCAN} (see \cref{alg:flow_est}, \cref{alg:line:clustering}) to form clusters that include points from both clouds.
Therefore, we select $n=8$ as it yields the lowest average estimation error while remaining applicable to the entire range of target velocities considered. 

\ale{Detailed results for $n=8$ are reported in \cref{tab:speed_comparison} for both the proposed approach and the baseline.}
The angular error in the velocity direction is computed at each time step $k$ and for each point as the angle between the ground truth and estimated velocity vectors:
\begin{equation*}
		\text{cos}^{-1} \frac{\bar{\mathbf{v}}_k \cdot \mathbf{v}_{i,k} }{\Vert \bar{\mathbf{v}}_k \Vert \Vert \mathbf{v}_{i,k} \Vert} .
\end{equation*}
Similarly to the previous case, it is averaged over time and across all points. \cref{tab:speed_comparison} reports the mean and standard deviation of this error across the five experimental trials for each target velocity.
\ale{Furthermore, the last two columns} of~\cref{tab:speed_comparison} report the corresponding statistics for the linear \gi{velocity error already used in the analysis of \cref{fig:res_n}.}
%

As visible from \cref{tab:speed_comparison}, for motion along the negative $y$-axis, \cref{alg:flow_est} accurately estimates the direction of the scene flow, with an average angular error below \SI{9}{\degree} across all four target speeds and with low standard deviation. Regarding the linear velocity error, \cref{alg:flow_est} estimates it with an average accuracy of approximately \SI{0.045}{\meter\per\second}. 
As shown in \cref{alg:flow_est}, the point-wise velocity is computed based on information on the displacement in the position, which is subject to uncertainty ranging from 8\% to 11\% as explained in \cref{sec:validation}. 
Since the point cloud is generated from raw measurements, it is reasonable to expect an error in the estimated velocity, which is coherent with the uncertainty of the measurements.
\ale{In contrast, the baseline method, which only applies clustering and ICP without the additional steps introduced in \cref{sec:methodology}, fails to accurately estimate both the motion direction and linear velocity. This highlights the necessity of the proposed modifications when working with low-resolution, noisy \gls{tof} data. 
} 

\ale{\cref{tab:speed_stats_others} instead reports the performance of \cref{alg:flow_est} in the remaining two experimental scenarios. As shown, both the direction and linear velocity estimation errors increase when the motion occurs along the $x$-axis.} 
In this case, the estimation is further degraded not only by the inherent sensor uncertainty but also by the increased sparsity of the measurements. Specifically, lateral motion (i.e., along the $x$-axis, sideways with respect to the robot) causes parts of the mannequin to fall outside the \gls{fov} of the \gls{tof} sensors, making it more difficult to establish consistent geometric correspondences between consecutive frames (see \cref{fig:res_flow}(c)). 
A similar trend is observed in the third set of experiments. Although the mannequin moves along the $y$-axis, the concurrent motion of the sensorised robot causes the \gls{fov} of each \gls{tof} to change over time, which adversely affects the accuracy of the scene flow estimation.

Finally, \cref{fig:res_flow} shows examples of the output of \cref{alg:flow_est} in both 3D and top views when considering the mannequin moving along both the $y$ and $x$ axes. 
%
%
Red points correspond to the starting point cloud $P_{k-n}$, while blue points represent $P_{k}$. The green arrows, associated with each red point, represent the estimated dense flow field; their magnitude has been scaled for clarity of visualization. It can be seen that the overall flow correctly points towards the direction of motion. Red points without associated arrows are those classified as stationary by the algorithm. \ale{Additionally, a supplementary video demonstrating the dense scene flow estimation for a human approaching the robot at three different speeds is provided with this paper.}

\section{Conclusions}

In this manuscript, we propose an algorithm to estimate the scene flow of moving objects in the surroundings of a robot using sparse and noisy point clouds obtained from miniaturised \gls{tof} sensors.

\ale{The algorithm builds upon ICP-based methods by incorporating additional filtering steps based on the ICP fitness score to improve robustness.}

For validation, we conducted experiments in three different scenarios involving two robotic manipulators -- one equipped with distributed \gls{tof} sensors and the other moving a mannequin.
The results demonstrate that the proposed method outperforms a baseline approach that does not include the additional filtering steps. 
Nevertheless, certain limitations remain, primarily due to the constraints of the hardware. Future work should focus on improving data processing to obtain more reliable measurements. Potential directions include applying distance correction techniques, such as those proposed in \cite{caroleo2025tiny}.
It is also worth investigating the use of geometric matching techniques better suited for point clouds that partially overlap, to address the aforementioned issue related to the sparsity.

The hierarchical clustering algorithm (\texttt{HDBSCAN}) employed may also pose limitations in more complex scenarios involving multiple humans and moving objects in the robot’s environment, potentially failing to cluster them accurately. Further analysis is needed to assess the robustness and effectiveness of the proposed pipeline under such conditions.

\bibliographystyle{IEEEtran} 
\bibliography{references.bib}


\end{document}